\documentclass[conference]{IEEEtran}
\IEEEoverridecommandlockouts
\usepackage{amsmath,amssymb,amsfonts}
\usepackage{amsthm}
\usepackage{booktabs}
\usepackage{multirow}
\usepackage{array}
\newcolumntype{P}[1]{>{\raggedright\arraybackslash}m{#1}}
\newcolumntype{C}[1]{>{\centering\arraybackslash}m{#1}}
\usepackage[table]{xcolor}
\usepackage{nicefrac}
\newcommand{\best}[1]{\textbf{#1}}
\newcommand{\ourscell}[1]{\cellcolor{gray!15}#1}
\usepackage{graphicx}
\usepackage{cite}
\usepackage{bbm}
\usepackage[hidelinks]{hyperref}
\usepackage{extpfeil}
\usepackage[normalem]{ulem}
\newcommand{\encodersep}{%
\specialrule{0.08em}{1pt}{0.08em}
\specialrule{0.08em}{0pt}{1pt}
}

\newtheorem{assumption}{Assumption}
\newtheorem{theorem}{Theorem}

\newcommand{\X}{\mathcal X}
\newcommand{\Ycal}{\mathcal Y}

\newcommand{\Hcal}{\mathcal H}
\newcommand{\Vcal}{\mathcal V}
\newcommand{\Dcb}{\mathcal D_{\mathrm{cb}}}
\newcommand{\Pc}{\Pr_{\Dcb}}

\newcommand{\botcls}{\bot}

\newcommand{\CE}{\operatorname{CE}}
\newcommand{\second}[1]{%
  \begingroup
  \renewcommand{\ULthickness}{0.45pt}%
  \setlength{\ULdepth}{1.0pt}%
  \uline{#1}%
  \endgroup
}
\usepackage{graphicx}
\usepackage{calc}

\newcommand{\smallpm}{%
\ensuremath{%
\mathbin{%
\vcenter{\hbox{%
\makebox[\widthof{$\pm$}][c]{\scalebox{0.8}{$\pm$}}%
}}%
}%
}%
}

\title{ProtoPur: Prototype-Purity Pseudo-Labeling for Weakly Supervised Instance Learning in Pathology
\thanks{* Corresponding Authors.}
}

\author{
\IEEEauthorblockN{
Chen Shu\textsuperscript{1},
Chengyang Zhang\textsuperscript{1,2},
Boyu Fu\textsuperscript{4},
Ting Yin\textsuperscript{2,3},
Wenchuan Zhang\textsuperscript{2,3},
Jie Chen\textsuperscript{2},
Yuhao Yi\textsuperscript{1,2,3 *},
Hong Bu\textsuperscript{2,3}
}

\IEEEauthorblockA{
\textsuperscript{1}\textit{College of Computer Science, Sichuan University, Chengdu, China}
}

\IEEEauthorblockA{
\textsuperscript{2}\textit{Institute of Clinical Pathology, West China Hospital, Sichuan University, Chengdu, China}
}

\IEEEauthorblockA{
\textsuperscript{3}\textit{Department of Pathology, West China Hospital, Sichuan University, Chengdu, China}
}

\IEEEauthorblockA{
\textsuperscript{4}\textit{Pittsburgh Institute, Sichuan University, Chengdu, China}
}

}

\begin{document}
\maketitle

\begin{abstract}
Weak bag-level labels are widely available in computational pathology, whereas reliable instance-level annotations for whole-slide images remain expensive and ambiguous. Existing weakly supervised multiple instance learning methods can achieve strong bag-level classification, but their instance-level supervision is often derived from attention scores, model predictions, negative bags, or spatial refinement, which may not reliably identify class-specific local evidence. We propose ProtoPur, a prototype-purity pseudo-labeling method for weakly supervised instance-level learning in pathology whole-slide images. ProtoPur estimates class-balanced local purity over prototype regions constructed from frozen foundation-model features, and assigns pseudo-labels to high-purity prototype regions that are strongly associated with a target class. This purity-first design separates class-specific evidence from negative or shared morphology, with coverage used only to ensure sufficient pseudo-label assignment.
We further provide a PAC-style analysis showing that the risk of an instance classifier trained with ProtoPur pseudo-labels can be decomposed into a standard uniform-convergence term and a ProtoPur-specific pseudo-label noise term, which consists of selected-region mismatch and a coverage residual. Experiments on three public whole-slide image datasets using two pathology foundation encoders show that ProtoPur consistently improves instance-level performance across multiple MIL backbones while maintaining competitive bag-level performance. Pseudo-label quality comparisons on two public datasets further demonstrate that prototype-purity selection provides stronger instance supervision. These results suggest that the foundation model-based features contain exploitable instance-level supervision signals, and ProtoPur is a way to reveal and use these signals from weak bag labels without requiring dense instance annotations. The code is available at \url{https://github.com/shuchen2024/ProtoPur}.
\end{abstract}


\section{Introduction}
Whole-slide images (WSIs) are the primary data format in computational pathology~\cite{campanella2019clinical,vanDerLaak2021,Cui2021}. They are gigapixel tissue images typically associated with a slide-level or specimen-level diagnosis. Since WSIs are too large to process directly, they are usually divided into patches and then formulated as multiple instance learning (MIL), where each slide is treated as a bag and patches are treated as instances. This formulation enables bag-level prediction under weak slide-level supervision, but many pathology tasks, such as cancer subtype classification, require not only accurate slide-level prediction but also localized patch-level evidence. Dense instance-level annotations are expensive and sometimes ambiguous, whereas slide-level or specimen-level labels are routinely available from pathology reports~\cite{campanella2019clinical,vanDerLaak2021}.

MIL has become a standard paradigm for weakly supervised WSI classification~\cite{abmil,clam,dsmil,transmil,DTFD,MHIM,DTMIL}. Attention-based~\cite{abmil,clam}, dual-stream~\cite{dsmil}, and Transformer-based~\cite{transmil} MIL models have achieved strong bag-level performance. However, bag-level accuracy does not necessarily imply reliable instance-level classification~\cite{INS,jang2024learnable}. A model may classify a slide correctly by relying on a small number of highly discriminative patches, while ignoring other relevant morphology or assigning high scores to non-specific regions. This mismatch is particularly important for subtyping, where the desired evidence should be specific to a disease category.

Recent weakly supervised methods have attempted to recover instance-level signals from slide-level labels by using attention scores, teacher predictions, negative bags, counterfactual interventions, or spatial refinement~\cite{weno,CIMIL,OT,smmile,INS,DTFD}. These strategies provide useful forms of instance supervision, but their pseudo-label sources are often tied to a trained bag model, a binary positive-negative assumption, or WSI-specific spatial structures. As a result, instance-level supervision can be sensitive to model bias, attention instability, shared morphology across classes, or the lack of explicit negative slides in subtype differentiation tasks.

A recent PAC formulation of instance learnability in MIL evaluates whether an instance-indexed predictor can predict the same observed label associated with the bag from an individual instance~\cite{jang2024learnable}. In that framework, the instance training sample pairs the $i$-th instance in each bag with the corresponding bag label, and the instance risk is defined with respect to the same label space as the bag risk. This is appropriate for analyzing whether an MIL architecture transfers bag-level supervision to instance-indexed predictors, but it does not address whether every instance should semantically inherit the bag label. 

In subtype classification tasks, 
some patches are negative, some correspond to shared or non-specific morphology, some are artifacts, and only a subset provides subtype-specific evidence. We therefore define a distinct instance-learning target: an instance receives subtype label $c$ only when its latent subtype-support set is the singleton $\{c\}$, while non-singleton or non-specific cases are assigned to a complement class. This formulation differs from classical MIL, which asks how instance labels aggregate into a bag label; our formulation asks which instances are entitled to a subtype label at all.

The key methodological observation is that weak labels are unreliable for individual instances but statistically informative over local semantic regions. For subtype classification, a prototype region that occurs almost exclusively in one subtype is a high-purity candidate for subtype-specific evidence, whereas a region that occurs broadly across subtypes is more likely to represent shared morphology. 
ProtoPur implements this idea by clustering instance embeddings into prototype regions, estimating class-balanced prototype purity from bag labels, and selecting high-purity regions before applying a mild coverage constraint. Our contributions are summarized as follows:

\begin{itemize}
\item We formulate weakly supervised instance-level learning for subtype classification, clarifying which instances are entitled to subtype labels. Based on the formulation, we provide a PAC-style analysis that separates classifier estimation error from ProtoPur-specific pseudo-label noise.

\item We propose ProtoPur, a scalable prototype-purity pseudo-labeling framework that estimates class-balanced local purity from weak bag labels and selects high-purity prototype regions as reliable instance evidence.

\item We conducted extensive experiments across multiple WSI cohorts, MIL backbones, and two pathology foundation encoders, demonstrating consistent instance-level improvements, stronger pseudo-label quality, and stable behavior over different prototype granularities.

\end{itemize}

\begin{figure*}[t]
\centering
\includegraphics[width=0.98\textwidth]{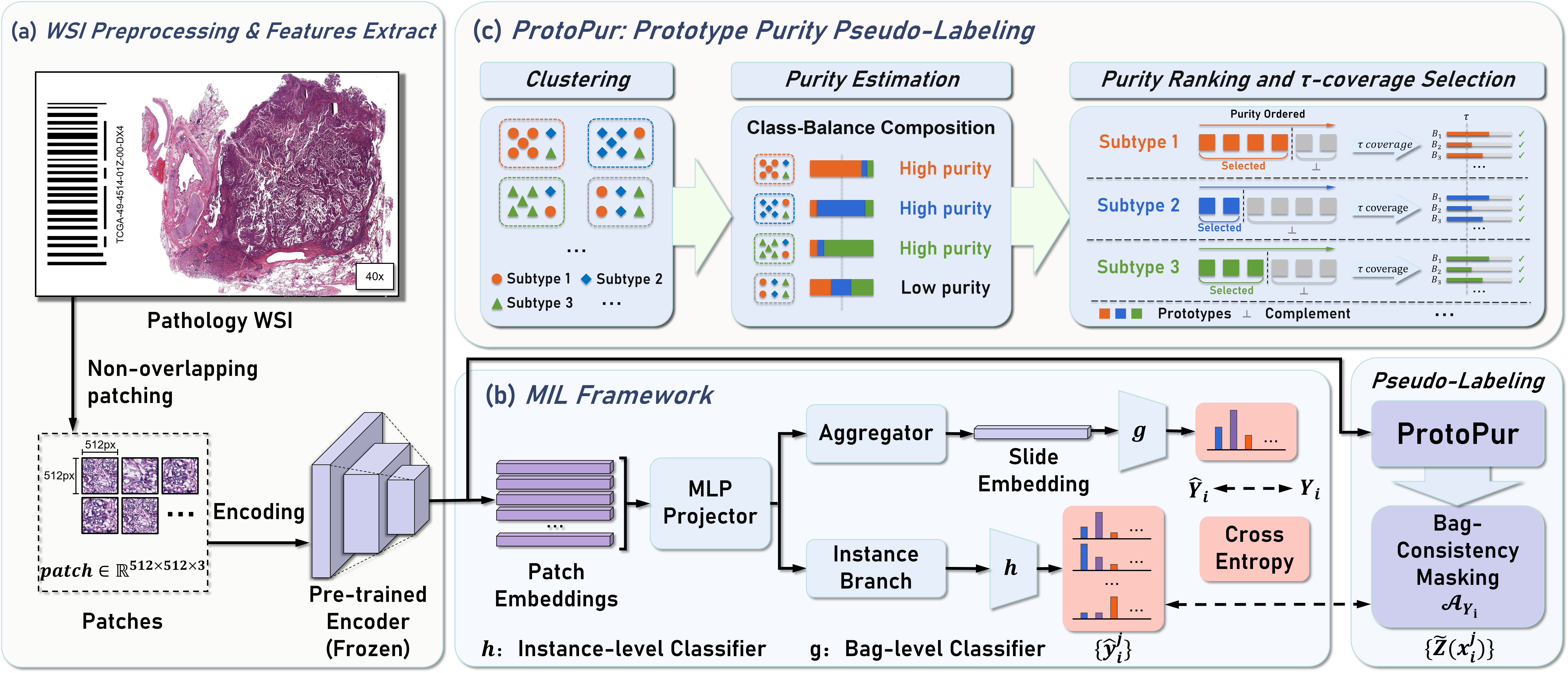}
\caption{Overview of ProtoPur as a weakly supervised instance pseudo-labeling framework integrated into an MIL architecture. (a) WSIs are divided into patches and encoded by a frozen pathology encoder. (b) The MIL framework includes both a bag branch and an instance branch for bag-level and instance-level prediction. (c) ProtoPur generates pseudo-labels by clustering instances, estimating prototype purity, and selecting high-purity prototypes under a $\tau$-coverage constraint. The pseudo-labels are filtered by the bag-consistency mask and used to supervise the instance branch of the MIL architecture.}
\label{fig:pipeline}
\end{figure*}

\section{Related Work}

\paragraph{Weakly supervised multiple instance learning}
MIL has been widely used for WSI classification because it can learn from slide-level labels without dense patch annotations. ABMIL introduces an attention-based aggregation mechanism that learns instance weights and produces a bag representation for classification~\cite{abmil}. CLAM extends attention-based MIL with clustering-constrained learning to improve data-efficient WSI classification~\cite{clam}. DSMIL uses a dual-stream design that combines instance-level classification with bag-level aggregation, allowing discriminative instances to contribute directly to slide prediction~\cite{dsmil}. TransMIL further models correlations among instances with a Transformer-based architecture, improving the representation of long-range contextual dependencies inside a slide~\cite{transmil}. These methods achieve strong bag-level performance, but their instance scores are mainly optimized to support bag classification and therefore do not necessarily correspond to reliable instance-level subtype labels.

\paragraph{Instance supervision from weak labels}
Several methods attempt to obtain instance-level supervision from weak WSI labels. 
CLAM uses attention-derived high-confidence instances to regularize the model through an auxiliary instance loss~\cite{clam}. 
WENO further distills teacher attention into instance-level supervision, transferring bag-model signals to patch-level training~\cite{weno}. 
CIMIL methods perturb or intervene on sub-bags to identify influential regions for bag prediction~\cite{CIMIL}. 
These approaches provide useful instance cues, but the generated pseudo-labels are still closely tied to the behavior of a trained bag model, attention mechanism, or negative-sample assumption. 
If the bag model relies on shortcut regions or unstable attention patterns, or if negative samples are unavailable in subtype-only tasks, the resulting instance supervision may become biased or less directly applicable.
PathPT explores vision-language prompt tuning for few-shot rare cancer subtyping, where text-side semantics is used to provide fine-grained patch-level guidance~\cite{pathpt}. However, the effectiveness of this approach is constrained by the zero-shot classification performance of vision-language models, making it better suited for few-shot learning than for scenarios with abundant bag-level supervision.

\paragraph{Self-training and spatial refinement}
SeLa-MIL formulates weakly supervised instance learning as a self-training problem in binary MIL, where instances from negative bags can be treated as reliable negative examples and positive instances are progressively mined from positive bags~\cite{OT}. This design is effective for positive-negative binary settings, but it is less directly applicable to subtype-only tasks where explicit negative bags may be unavailable and each WSI belongs to one target subtype. SMMILe improves WSI spatial quantification by combining instance scores, superpatch construction, refinement networks, and Markov random field smoothing~\cite{smmile}. Its spatial refinement is valuable for WSI-level localization, 
but it relies on WSI-specific spatial structures and does not fully leverage the semantic locality of instances across slides.

The proposed ProtoPur is complementary to the MIL backbones and weak supervision strategies. Instead of deriving pseudo-labels from attention scores, teacher predictions, or negative bags, ProtoPur estimates class-balanced occurrence purity at the prototype-level. A prototype region is selected as class-specific evidence only when it is locally pure for one class under weak bag labels. This purity-first design separates class-specific evidence from shared morphology and uses coverage only as a sufficiency constraint. Therefore, ProtoPur can be integrated with different MIL backbones and feature encoders to provide more reliable instance-level supervision without being tied to a particular bag-model output or WSI-grid-based refinement procedure.

\section{Method}

\subsection{Problem Formulation}
In the weakly supervised MIL setting, each WSI is treated as a bag, and its patches are treated as instances. The training set is given as a weakly labeled set $S:=\{(B_i,Y_i)\}_{i=1}^N$, where each bag $B_i:=\{x_i^1, x_i^2, \ldots, x_i^{n_i}\}$ is associated with an observed bag-level label $Y_i$, and the instance-level labels are not observed.

Let $\mathcal C:=\{1,2,\ldots, C\}$ denote the set of target positive concepts, such as disease subtypes, and let $\Ycal$ denote the observed bag-label space. In the classification of multiple-subtype WSIs without pure negative samples, $\Ycal=\mathcal C$, and in classical positive-negative MIL under the Boolean OR assumption~\cite{blum1998note,sabato2012multi,doran2016multiple}, $\mathcal C=\{1\}$ and $\Ycal=\{0,1\}$, where the label $0$ denotes the negative class.

We further define $\Dcb$ as the class-balanced instance distribution induced by the following sampling scheme: (1) sample a class uniformly from $\mathcal Y$; (2) sample a bag conditional on that class; (3) sample an instance from the selected bag. This distribution is the population analog of the class-balanced occurrence statistics used by ProtoPur, where the number of instances assigned to each prototype is normalized by the number of bags per class. Therefore, classes with a larger number of bags do not dominate the purity estimate.

\paragraph{Singleton-support instance label}
Let $\mathcal X$ be the instance space. Each $x\in\mathcal X$ has a latent target-concept support set $\mathcal S(x)\subseteq\mathcal C$, where $c\in\mathcal S(x)$ means that the semantic pattern of $x$ appears with non-negligible probability in bags of target class $c$. The singleton-support label of $x$ is defined as
\[
Z(x):=
\begin{cases}
c, & \mathcal S(x)=\{c\},\\
\botcls, & |\mathcal S(x)|\neq 1.
\end{cases}
\]
The complement $\botcls$ is an operational heterogeneous class for instances without a target subtype label, covering true negatives, shared or non-specific morphology, and artifacts. The goal is to learn $h:\X\to\mathcal C\cup\{\botcls\}$ with low singleton-support risk $$ R_Z(h)= \Pr_{X\sim \mathcal D_{\mathrm{cb}}}\big(h(X)\neq Z(X)\big). $$

\subsection{Semantic locality and prototype partition}
A pretrained pathology foundation model $\phi$ is used to encode each instance $x$ into an embedding $z=\phi(x)$~\cite{conch,uni}. ProtoPur clusters instances from all training bags in the embedding space using $k$-means, obtaining a prototype partition $\Vcal=\{V_l\}_{l=1}^L$. Each $V_l$ is a local semantic region. ProtoPur is encoder-agnostic in implementation but representation-dependent in performance; a stronger encoder should make subtype-specific singleton evidence more compact and class-pure, while separating it from shared morphology and patterns.

ProtoPur exploits locality at the prototype level: nearby instances in the feature space are aggregated into prototypes, whose class associations are then estimated collectively rather than from point-wise instance scores. Weak labels are noisy for individual instances, but they become statistically informative when pooled over semantic regions. The prototypes provide a stable and scalable region family for class-balanced purity estimation and for subsequent coverage-controlled selection.

\subsection{Purity-first selection with \texorpdfstring{$\tau$}{tau}-coverage}

\paragraph{Bag-normalized prototype purity} For a prototype $V_l$ and a target concept $c\in\mathcal C$, let $m_{c,l}$ be the number of instances in $V_l$ from label-$c$ bags, and let $N_c$ be the number of label-$c$ bags. Define $\widetilde m_{c,l}=\nicefrac{m_{c,l}}{N_c}$. ProtoPur estimates task-specific local purity by
$\widehat{q}_{l,c}=\nicefrac{\widetilde m_{c,l}}{(\sum_{c'\in\mathcal C}\widetilde m_{c',l})}$.
 Normalization of the number of bags prevents labels with more bags from dominating the purity estimates.

\paragraph{Pseudo-labeling} For subtype classification tasks, ProtoPur ranks prototypes for each $c$ in decreasing order of $\widehat{q}_{l,c}$. Starting from the most class-pure prototypes, it adds prototypes to a set $T_c(\tau)$ until every eligible positive bag for concept $c$ contains at least $\tau$ selected instances. If a prototype is selected by multiple targets, it is assigned to the target with the largest purity score. We define
$I_c(\tau)=\bigcup_{l \in T_c(\tau)}V_l$, $I(\tau)=\bigcup_{c\in\mathcal C}I_c(\tau)$,
and use $\mathcal A_{Y_i}\subseteq \mathcal C \cup \{\botcls\}$ to denote the set of instance labels that are compatible with the bag label $Y_i$. ProtoPur assigns a pseudo-label to an instance as follows:
\[
\tilde Z(x_i^j)=
\begin{cases}
\botcls, & Y_i = 0\lor x_i^j\notin I(\tau)\,,\\
c, & Y_i\neq 0 \land x_i^j\in I_c(\tau) \land c \in \mathcal A_{Y_i},\\
\text{unassigned}, &\text{otherwise}.
\end{cases}
\]
Selection stops when the least-covered eligible bag
contains $\tau$ selected instances, so labeled instance counts may vary across bags.
The compatibility mask \(\mathcal A_{Y_i}\) prevents pseudo-labels that
contradict the bag label; incompatible selected instances are ignored. Thus,
ProtoPur prioritizes pseudo-label reliability over exhaustive coverage. $Y_i=0$ denotes a pure negative bag, which may be absent in subtype-only cohorts.

\paragraph{Using ProtoPur inside MIL}
ProtoPur is a pseudo-labeling method rather than a specific MIL architecture. The pseudo-labels $\tilde{Z}$ produced by ProtoPur can be integrated into an MIL model. The default instantiation used in this paper is an auxiliary instance loss. In implementation, this loss is the usual cross-entropy surrogate. Instances assigned to \(\botcls\) supervise the complement class of the
instance classifier, while unassigned instances are excluded from the
instance-level loss. Let $g$ and $h$ denote a bag classifier and an instance classifier, respectively. The training objective is
\[
\widehat{\mathcal L}^{\mathrm{CE}}=
\widehat{\mathcal L}_{\mathrm{bag}}^{\mathrm{CE}}\big(g;Y\big)+
\lambda\widehat{\mathcal L}_{\mathrm{ins}}^{\mathrm{CE}}\big(h;\tilde{Z}\big),
\]
where $\widehat{\mathcal L}_{\mathrm{bag}}^{\mathrm{CE}}(g;Y)$ is defined as the average of $\CE\bigl(Y_i,g(B_i)\bigr)$ over all bags, and $\widehat{\mathcal L}^{\mathrm{CE}}_{\mathrm{ins}}(h;\tilde Z)$ is defined as the average of $\mathrm{CE}\big(\tilde Z(x_i^j),h(x_i^j)\big)$ over instances with assigned pseudo-labels. 

\section{Theoretical Analysis}
This section analyzes the learnability of the proposed method. We provide a proof sketch for the main theorem.
\subsection{Pseudo-label error decomposition}
ProtoPur generates the pseudo-label $\tilde{Z}$. Its error with respect to the latent singleton-support label is
$\rho=\Pr_{\mathcal D_{\mathrm{cb}}}(\tilde{Z}(X)\neq Z(X))$. 
This error decomposes into a coverage residual and a selected-region impurity:
\[
\begin{aligned}
\rho
=&\Pc\big(\tilde{Z}(X)\in\mathcal C,\tilde{Z}(X)\neq Z(X)\big) \\
&+\Pc\big(Z(X)\in\mathcal C,\tilde{Z}(X)=\botcls\big).
\end{aligned}
\]
The first term, denoted by $\xi$, measures the error introduced when selected regions contain instances that do not match their assigned pseudo-labels. The second term, denoted by $\beta$,  accounts for true singleton-support evidence that is not assigned pseudo-labels by ProtoPur, because its selection rule prioritizes high-purity prototypes rather than exhaustively labeling all potentially relevant regions. In this paper, we keep $\beta$ as an explicit residual and evaluate its practical effect through $\tau$-sensitivity and pseudo-label coverage, rather than claiming exhaustive recovery of all singleton-support evidence.

\subsection{One-vs-rest exposure}
For subtype $c$, ProtoPur can directly assess whether the selected regions come from bags labeled $c$, but it cannot observe the latent singleton-support label of each selected instance. We therefore define the selected-region bag-label mismatch rate as
\begin{align}
\alpha_{c,\tau}:=&
\Pr\limits_{X\sim \mathcal D_{\mathrm{cb}}}
\big(Y\neq c, \mid X\in I_c(\tau)\big)\nonumber\\
=&\Pr\limits_{X\sim \mathcal D_{\mathrm{cb}}}
\big(Y\neq c, Z(x)\neq c\mid X\in I_c(\tau)\big).\nonumber
\end{align}
The equality holds since $Y\neq c \Rightarrow Z(x)\neq c, \forall c\in \mathcal C$. This quantity measures how often an instance selected for the subtype $c$ comes from a bag whose label is not $c$. The following exposure condition relates this observable bag-level mismatch rate to the latent singleton-level pseudo-label error.

\begin{assumption}[One-vs-rest exposure]
For each subtype $c$, singleton subtype-$c$ instances appear only in class-$c$ bags, and there exists $\gamma_c>0$ for all selected regions $I(\tau)$,
\[
\Pr\limits_{X\sim \mathcal D_{\mathrm{cb}}}
\big(Y\neq c\mid X \in I(\tau), Z(X)\neq c\big)\ge \gamma_c.
\]
\end{assumption}
Intuitively, if an instance is not subtype-$c$ singleton evidence, it must have nonzero exposure in non-$c$ bags. Otherwise, a subtype-associated non-specific pattern would be observationally indistinguishable from true singleton-support evidence. Under this assumption, with probability at least $1-\delta_\alpha$, we attain
\[
\Pr\limits_{X\sim \mathcal D_{\mathrm{cb}}}
\big(Z(X)\neq c\mid X\in I_c(\tau)\big)
\le
\frac{\hat\alpha_{c,\tau}+\varepsilon_n}{\gamma_c},
\]
in which $\hat\alpha_{c,\tau}$ denotes the empirical fraction of selected instances in $I_c(\tau)$ whose bag label is not $c$, and $\varepsilon_n$ accounts for finite-sample estimation error. Thus, the latent singleton-level error of the selected subtype-$c$ region is bounded by the observable bag-label mismatch rate, scaled by the exposure factor $\gamma_c$.

\subsection{Main learnability statement}
Define the observable pseudo-label noise upper bound
\[
\rho^{up}=
\beta+
\sum_{c\in\mathcal C}
\Pr\limits_{X\sim \mathcal D_{\mathrm{cb}}}
\big(X\in I_c(\tau)\big)
\frac{\hat\alpha_{c,\tau}+\varepsilon_n}{\gamma_c}.
\]

\begin{theorem}[ProtoPur-induced singleton-support instance learnability]
\label{thm:main}
After pseudo-labeling by ProtoPur, the selected regions $I(\tau)$ are disjoint. Given the one-vs-rest exposure condition and the finite-sample bag-label mismatch estimation event, let $\hat h$ be an empirical risk minimizer over $\Hcal$ trained on ProtoPur pseudo-labels. If the pseudo-label risk satisfies a uniform convergence bound $\mathcal E_n(\Hcal,\delta_H)$, then with probability at least $1-\delta_\alpha-\delta_H$,
\[
R_Z(\widehat h)
\le
\inf_{h\in\Hcal}R_Z(h)
+2\mathcal E_n(\Hcal,\delta_H)
+2\rho^{up}.
\]
\end{theorem}

The risks in Theorem~1 are $0$-$1$ classification risks used for learnability analysis. 
The selected-region impurity term controls pseudo-label precision; $\beta$ is a coverage residual that reflects the fact that ProtoPur intentionally abstains from ambiguous or lower-purity regions.

Theorem~1 separates the singleton-support instance risk into three terms: the best achievable risk within $\Hcal$, the standard estimation error $\mathcal E_n(\Hcal,\delta_H)$, and the ProtoPur-induced pseudo-label noise $\rho^{up}$. The estimation term can be instantiated by standard uniform-convergence bounds for the induced $0$-$1$ loss class~\cite{vapnik1998statistical,mohri2018foundations,shalev2014understanding}. Let $d_{\Hcal}$ be the Vapnik–Chervonenkis (VC) dimension of the $0$-$1$ loss class induced by the instance hypothesis class $\Hcal$, and $\eta:=\nicefrac{\max_i n_i}{
min_i n_i}$ be the maximum ratio of the numbers of instances of two bags. Then
$
\mathcal E_n(\Hcal,\delta_H)=
O(
(
\nicefrac{(d_{\Hcal}\log \widetilde{N}+\log(1/\delta_H))}
{\widetilde{N}}
)^{1/2})$, 
where $\widetilde{N}:= \nicefrac{N}{\eta}$. 
This term reflects the capacity and sample size of the instance classifier. In contrast, $\rho^{up}$ is specific to ProtoPur and is controlled by selected-region mismatch, coverage residual, and finite-sample purity estimation. Better encoders can reduce this pseudo-labeling term by forming more compact and class-pure prototype regions, making high-purity prototypes easier to rank and select. Thus, ProtoPur does not reduce the intrinsic capacity of the neural classifier; rather, it reduces the pseudo-label noise induced by weak supervision.

\section{Experiments}

\subsection{Experimental Setting}
\label{sec:exp_setting}
\paragraph{Datasets}
We evaluated the proposed ProtoPur in the main bag-level and instance-level classification experiments on three publicly available datasets: TCGA-BRCA-Invasive~\cite{TCGA_BRCA}, TCGA-LUNG~\cite{TCGA_LUAD,TCGA_LUSC}, and BRACS~\cite{BRACS}. TCGA-BRCA-Invasive is constructed from the invasive carcinoma subset of TCGA-BRCA rather than the entire TCGA-BRCA cohort, and is used to assess multiple-subtype classification of invasive breast cancer WSIs. TCGA-LUNG includes lung adenocarcinoma and lung squamous cell carcinoma WSIs, forming a lung cancer subtype classification task. BRACS is a breast pathology dataset with region-level annotations. 
TCGA-BRCA-Invasive and TCGA-LUNG are subtype-only classification tasks, where each WSI belongs to one of the target subtypes, whereas BRACS additionally contains negative samples, resulting in a classification setting with both target lesion categories and negative cases. 
For TCGA-BRCA-Invasive and TCGA-LUNG, the instance-level evaluation labels are derived from public manual tumor-tissue annotations of TCGA diagnostic WSIs~\cite{loeffler2021manualtumor}, which were originally created to enrich invasive tumor regions for molecular prediction studies rather than to provide exhaustive pixel-level tumor segmentation. Patches overlapping annotated tumor regions are treated as foreground tumor instances and inherit the corresponding slide subtype, whereas patches outside annotated tumor regions are treated as background. For BRACS, the available region-level annotations are used in the same evaluation-only manner, while the unannotated regions are excluded from evaluation. Note that these annotations delineate tumor regions rather than subtype-specific singleton evidence, so they serve only as a proxy for $Z$. TCGA-PanGlioma~\cite{TCGA_PanGlioma} is additionally used for pseudo-label quality evaluation (Section~\ref{sec:label_quality}). 

\paragraph{Evaluation metrics}
For bag-level performance, we report bag-level macro AUC. For instance-level evaluation, we report three complementary AUC metrics: $\mathrm{AUC}_{\mathcal Y}$, $\mathrm{AUC}_{\mathcal C}$ and $\mathrm{AUC}_{\mathcal C\mid FG}$. $\mathrm{AUC}_{\mathcal Y}$ is the macro-average AUC over the full instance label space $\mathcal Y$, calculated using the prediction probabilities of all instances for all instance classes, including both subtype classes and the negative class. $\mathrm{AUC}_{\mathcal C}$ is the macro-average AUC over the subtype classes, and it is also calculated using the subtype prediction probabilities for all instances, including negative instances. In this metric, the negative class is not included as an averaged target class, but negative instances are retained in the evaluation. $\mathrm{AUC}_{\mathcal C\mid FG}$ is calculated over foreground instances only, after removing negative instances, and therefore focuses on subtype-to-subtype discrimination.
These metrics provide a fine-grained evaluation of instance-level classification. $\mathrm{AUC}_{\mathcal Y}$ measures overall discrimination across the full label space, $\mathrm{AUC}_{\mathcal C}$ evaluates subtype scores in the presence of negative instances, and $\mathrm{AUC}_{\mathcal C\mid FG}$ isolates subtype discrimination among foreground instances. 

\paragraph{Implementation Details}
All experiments were conducted on NVIDIA RTX 4090 GPUs. For preprocessing, each WSI was tiled into non-overlapping $512\times512$ patches at $40\times$ magnification. Patch features were extracted offline using frozen CONCH~\cite{conch} and UNI~\cite{uni} encoders, yielding 512-dimensional and 1024-dimensional embeddings, respectively, and all subsequent models were trained on these features.
For training and evaluation, we adopted a $4$-fold cross-validation protocol following common MIL practice. In each fold, the data was split at the WSI level into training, validation, and test sets with a ratio of 2:1:1. The validation split was used for early stopping and model selection, and the held-out test set was used only for final evaluation. Hyperparameters were tuned on the validation splits only and then held fixed across all datasets, backbones, and encoders, so that no test data influenced any hyperparameter choice. All methods share the same folds and random seed. 
All models were trained using the Adam optimizer with a learning rate of $2\times10^{-4}$ and a weight decay of $1\times10^{-5}$, and the bag-level batch size was set to 1. 
The maximum number of training epochs was set to $100$, and early stopping was applied with a patience of 20 epochs.
For ProtoPur, we used $k$-means for clustering. The number of prototypes was set to $k=2000$ and the coverage threshold to $\tau=5$ for both CONCH and UNI features. The auxiliary instance loss weight $\lambda$ was set to $0.3$ for all experiments. 
All results in the main tables use this single shared hyperparameter configuration, while sensitivity analysis is given in Section \ref{sec:ablation}.
For comparison methods, WENO~\cite{weno}, INS~\cite{INS}, and SeLa-MIL~\cite{OT} were originally designed for positive-negative binary classification at the instance level. 
To adapt them to the multiple-subtype setting, we used CLAM-style one-vs-rest binary instance classifiers for each subtype~\cite{clam}.
Since the official SeLa-MIL implementation was unavailable, we reproduced it based on the original paper and used one-vs-rest negative-class construction during training. 
We integrated ProtoPur with DS-MIL, CLAM, and SMMILe as a plug-and-play pseudo-labeling module, since these backbones contain instance-level branches for direct pseudo-label supervision.

\subsection{Results and comparisons}
\begin{table*}[t]
\centering
\caption{Results on three cohorts with CONCH and UNI encoders ($k=2000$ for both). All AUC values are reported as percentages. Best and second-best results within each encoder are highlighted in bold and underlined, respectively.}
\label{tab:main_all}
\begingroup
\tiny
\renewcommand{\arraystretch}{0.82}
\setlength{\tabcolsep}{1.45pt}
\resizebox{\textwidth}{!}{%
\begin{tabular}{@{}cccccccccccccc@{}}
\toprule
\multirow{3}{*}{\raisebox{-2.0ex}{Encoder}} 
& \multirow{3}{*}{\raisebox{-2.0ex}{Model}}
& \multicolumn{4}{c}{TCGA-BRCA-Invasive}
& \multicolumn{4}{c}{TCGA-LUNG}
& \multicolumn{4}{c}{BRACS} \\
\cmidrule(lr){3-6}\cmidrule(lr){7-10}\cmidrule(lr){11-14}
& &
\multicolumn{1}{c}{Bag} & \multicolumn{3}{c}{Instance}
& \multicolumn{1}{c}{Bag} & \multicolumn{3}{c}{Instance}
& \multicolumn{1}{c}{Bag} & \multicolumn{3}{c}{Instance} \\
\cmidrule(lr){3-3}\cmidrule(lr){4-6}
\cmidrule(lr){7-7}\cmidrule(lr){8-10}
\cmidrule(lr){11-11}\cmidrule(lr){12-14}
& &
AUC & $\mathrm{AUC}_{\mathcal Y}$ & $\mathrm{AUC}_{\mathcal C}$ & $\mathrm{AUC}_{\mathcal C\mid FG}$
& AUC & $\mathrm{AUC}_{\mathcal Y}$ & $\mathrm{AUC}_{\mathcal C}$ & $\mathrm{AUC}_{\mathcal C\mid FG}$
& AUC & $\mathrm{AUC}_{\mathcal Y}$ & $\mathrm{AUC}_{\mathcal C}$ & $\mathrm{AUC}_{\mathcal C\mid FG}$ \\
\midrule

\multirow{12}{*}{CONCH~\cite{conch}}
& AB-MIL~\cite{abmil}
& 88.41\smallpm2.75 & N/A & N/A & N/A
& \second{98.31\smallpm0.50} & N/A & N/A & N/A
& 84.63\smallpm2.16 & N/A & N/A & N/A \\

& Trans-MIL~\cite{transmil}
& 86.51\smallpm2.93 & N/A & N/A & N/A
& 97.71\smallpm0.44 & N/A & N/A & N/A
& 82.70\smallpm1.23 & N/A & N/A & N/A \\

& DTFD-MIL~\cite{DTFD}
& 87.11\smallpm2.87 & 71.70\smallpm1.49 & 73.36\smallpm1.47 & 77.86\smallpm2.50
& 98.21\smallpm0.56 & 72.59\smallpm1.08 & 76.49\smallpm0.99 & 84.46\smallpm0.92
& 82.58\smallpm1.32 & 80.07\smallpm1.52 & 80.92\smallpm1.40 & 82.64\smallpm1.53 \\

& WENO~\cite{weno}
& 86.75\smallpm4.57 & 79.74\smallpm0.90 & 77.63\smallpm1.96 & 79.09\smallpm3.53
& 98.19\smallpm0.52 & 78.57\smallpm1.35 & 79.56\smallpm1.07 & 84.30\smallpm1.37
& 84.37\smallpm1.46 & 76.64\smallpm2.88 & 78.03\smallpm3.56 & 80.53\smallpm3.35 \\

& SeLa-MIL~\cite{OT}
& 87.66\smallpm2.45 & 82.73\smallpm1.32 & 81.27\smallpm1.44 & 78.18\smallpm1.89
& 97.04\smallpm1.30 & 68.46\smallpm1.50 & 67.27\smallpm1.14 & 77.33\smallpm2.31
& 79.69\smallpm0.65 & 74.52\smallpm2.86 & 76.90\smallpm2.88 & 77.21\smallpm3.08 \\

& INS~\cite{INS}
& \best{89.55\smallpm1.80} & 75.13\smallpm11.9 & 76.70\smallpm8.04 & 79.39\smallpm2.34
& 97.83\smallpm0.57 & 62.72\smallpm4.36 & 75.78\smallpm3.25 & 88.26\smallpm0.97
& 79.51\smallpm1.81 & 77.43\smallpm2.15 & 78.54\smallpm1.08 & 80.20\smallpm1.91 \\

& DS-MIL~\cite{dsmil}
& 88.47\smallpm3.27 & 72.83\smallpm3.10 & 71.30\smallpm3.47 & 78.62\smallpm2.89
& 97.67\smallpm0.80 & 71.50\smallpm1.80 & 75.31\smallpm0.89 & 83.06\smallpm1.03
& 83.76\smallpm1.58 & 56.89\smallpm2.58 & 59.44\smallpm2.76 & 59.19\smallpm2.54 \\

& CLAM~\cite{clam}
& 87.23\smallpm3.31 & 83.51\smallpm0.57 & 81.15\smallpm0.67 & 77.16\smallpm1.96
& 98.29\smallpm0.63 & 77.77\smallpm0.95 & 77.21\smallpm0.53 & 79.42\smallpm2.67
& 84.46\smallpm1.96 & 80.49\smallpm2.16 & 80.06\smallpm2.62 & 80.85\smallpm2.52 \\

& SMMILe~\cite{smmile}
& 86.68\smallpm3.76 & \second{86.25\smallpm2.38} & \second{83.98\smallpm3.10} & \second{81.11\smallpm3.32}
& 97.78\smallpm0.66 & \second{87.25\smallpm0.41} & \second{87.79\smallpm0.44} & \second{91.43\smallpm0.59}
& 76.58\smallpm3.15 & 78.58\smallpm2.30 & 78.43\smallpm2.60 & 81.15\smallpm2.64 \\

\cmidrule[0.08em](l){2-14}

& DS-MIL+Ours
& \second{89.14\smallpm1.40} & 82.20\smallpm1.75 & 79.75\smallpm2.86 & 78.70\smallpm3.41
& 97.84\smallpm0.68 & 77.66\smallpm2.18 & 79.04\smallpm1.60 & 85.38\smallpm1.25
& \best{85.31\smallpm2.73} & 75.82\smallpm1.83 & 77.16\smallpm1.86 & 78.81\smallpm2.04 \\

& CLAM+Ours
& 88.39\smallpm2.59 & 84.76\smallpm1.34 & 83.43\smallpm1.86 & 79.76\smallpm2.36
& \best{98.42\smallpm0.62} & 78.04\smallpm1.28 & 79.91\smallpm1.15 & 87.83\smallpm0.53
& \second{84.77\smallpm2.39} & \best{81.59\smallpm1.51} & \best{81.92\smallpm1.55} & \second{83.09\smallpm1.87} \\

& SMMILe+Ours
& 88.16\smallpm3.76 & \best{87.19\smallpm1.84} & \best{85.28\smallpm2.51} & \best{82.86\smallpm3.49}
& 97.93\smallpm0.56 & \best{87.53\smallpm0.64} & \best{88.25\smallpm0.77} & \best{91.50\smallpm0.66}
& 80.45\smallpm1.78 & \second{81.34\smallpm1.96} & \second{81.45\smallpm2.28} & \best{83.43\smallpm2.31} \\

\encodersep

\multirow{12}{*}{UNI~\cite{uni}}
& AB-MIL~\cite{abmil}
& 86.73\smallpm6.18 & N/A & N/A & N/A
& 98.43\smallpm0.39 & N/A & N/A & N/A
& 81.98\smallpm1.81 & N/A & N/A & N/A \\

& Trans-MIL~\cite{transmil}
& 85.92\smallpm5.59 & N/A & N/A & N/A
& 98.21\smallpm0.48 & N/A & N/A & N/A
& 80.44\smallpm2.25 & N/A & N/A & N/A \\

& DTFD-MIL~\cite{DTFD}
& \second{89.15\smallpm4.01} & 70.08\smallpm3.60 & 73.99\smallpm4.93 & 80.32\smallpm5.29
& 98.22\smallpm0.63 & 76.28\smallpm2.82 & 80.27\smallpm1.69 & 88.98\smallpm1.05
& 82.11\smallpm3.24 & 78.06\smallpm3.24 & 80.35\smallpm3.66 & 82.51\smallpm3.37 \\

& WENO~\cite{weno}
& 87.79\smallpm6.58 & 76.55\smallpm5.28 & 75.64\smallpm5.23 & 75.41\smallpm5.67
& 98.45\smallpm0.66 & 78.56\smallpm1.82 & 81.94\smallpm0.34 & 89.04\smallpm1.07
& \best{83.48\smallpm0.44} & 79.00\smallpm1.92 & 80.81\smallpm1.45 & 83.05\smallpm1.28 \\

& SeLa-MIL~\cite{OT}
& 86.71\smallpm4.84 & 69.65\smallpm10.3 & 71.46\smallpm5.68 & 74.10\smallpm1.80
& 97.17\smallpm0.91 & 77.09\smallpm5.44 & 79.66\smallpm5.76 & 90.74\smallpm1.85
& 77.91\smallpm2.22 & 78.13\smallpm2.53 & 79.78\smallpm1.79 & 81.51\smallpm1.85 \\

& INS~\cite{INS}
& 88.64\smallpm7.41 & 73.03\smallpm4.58 & 75.53\smallpm3.59 & 80.41\smallpm4.09
& \best{98.56\smallpm0.31} & 72.92\smallpm1.97 & 82.72\smallpm1.62 & \second{93.16\smallpm1.31}
& 76.57\smallpm5.54 & 74.53\smallpm2.80 & 76.22\smallpm2.35 & 77.16\smallpm2.43 \\

& DS-MIL~\cite{dsmil}
& 87.81\smallpm5.77 & 60.65\smallpm3.50 & 66.71\smallpm3.58 & 74.60\smallpm4.35
& 98.07\smallpm0.68 & 72.34\smallpm2.02 & 76.65\smallpm1.85 & 85.41\smallpm1.76
& 81.55\smallpm3.99 & 72.88\smallpm2.39 & 72.94\smallpm2.30 & 74.39\smallpm2.27 \\

& CLAM~\cite{clam}
& 88.35\smallpm6.32 & 81.07\smallpm2.02 & 79.18\smallpm3.18 & 78.88\smallpm3.75
& 98.35\smallpm0.32 & 77.78\smallpm1.57 & 78.89\smallpm1.38 & 87.52\smallpm1.02
& 82.36\smallpm3.18 & 80.30\smallpm2.10 & 79.62\smallpm1.89 & 80.55\smallpm2.13 \\

& SMMILe~\cite{smmile}
& 87.87\smallpm5.22 & \second{86.53\smallpm2.69} & \second{84.39\smallpm3.49} & \second{82.26\smallpm4.66}
& 98.25\smallpm0.55 & \second{87.60\smallpm1.09} & \second{88.49\smallpm1.25} & 93.04\smallpm0.89
& 77.97\smallpm4.50 & 78.19\smallpm3.14 & 79.04\smallpm2.56 & 81.61\smallpm2.32 \\

\cmidrule[0.08em](l){2-14}

& DS-MIL+Ours
& 88.29\smallpm3.92 & 80.50\smallpm2.61 & 79.22\smallpm3.81 & 79.22\smallpm4.15
& 98.26\smallpm0.55 & 78.92\smallpm0.71 & 81.79\smallpm0.40 & 90.05\smallpm0.69
& 81.92\smallpm3.17 & 78.44\smallpm3.54 & 79.69\smallpm2.88 & 81.36\smallpm2.83 \\

& CLAM+Ours
& 89.01\smallpm6.06 & 83.62\smallpm2.62 & 82.83\smallpm2.59 & 80.67\smallpm3.74
& 98.52\smallpm0.31 & 79.51\smallpm0.70 & 82.44\smallpm0.83 & 91.83\smallpm0.92
& \second{82.56\smallpm3.28} & \best{82.68\smallpm2.17} & \best{82.37\smallpm2.21} & \best{83.39\smallpm2.20} \\

& SMMILe+Ours
& \best{89.64\smallpm4.61} & \best{88.65\smallpm2.25} & \best{86.99\smallpm2.91} & \best{83.53\smallpm3.51}
& \second{98.53\smallpm0.44} & \best{87.82\smallpm0.69} & \best{88.69\smallpm0.75} & \best{93.24\smallpm1.23}
& 79.67\smallpm3.49 & \second{80.89\smallpm4.13} & \second{81.15\smallpm4.28} & \second{83.24\smallpm4.07} \\

\bottomrule
\end{tabular}%
}
\endgroup
\end{table*}

Table~\ref{tab:main_all} compares ProtoPur-enhanced MIL models with representative weakly supervised MIL baselines on three datasets under both bag-level and instance-level evaluation. Instance metrics use only instance classifier outputs. Overall, incorporating ProtoPur consistently improves instance-level classification across different datasets, MIL backbones, and feature encoders. The gains appear not only in $\mathrm{AUC}_{\mathcal Y}$ but also in the subtype-related metrics $\mathrm{AUC}_{\mathcal C}$ and $\mathrm{AUC}_{\mathcal C\mid FG}$, suggesting that ProtoPur provides useful instance-level supervision for learning subtype-discriminative evidence rather than only separating tumor and negative regions.
Across the three datasets, ProtoPur improves several representative MIL architectures, including DS-MIL, CLAM, and SMMILe. Compared with their original versions, the ProtoPur-enhanced variants generally achieve stronger instance-level AUCs while maintaining competitive bag-level performance. This trend is especially important because many MIL models can obtain high bag-level AUCs without producing reliable instance-level predictions. By introducing prototype-based pseudo-labels, ProtoPur helps the instance classifier learn more subtype-aware local representations, leading to more consistent gains in instance-level evaluation.
The results also show that the benefits of ProtoPur are not limited to a specific MIL backbone or dataset type. On TCGA-BRCA-Invasive and TCGA-LUNG, where the task focuses on subtype-only classification, ProtoPur improves subtype-level instance discrimination. On BRACS, which additionally contains negative samples, ProtoPur remains effective under a more challenging setting involving both lesion categories and negative instances. These results indicate that ProtoPur can improve instance-level learning in both subtype-only and negative-included classification scenarios.
ProtoPur works with both CONCH and UNI features; the consistent improvements under both encoders suggest that the proposed method is not tied to a specific type of feature representation. Instead, ProtoPur can provide effective instance-level supervision for features encoded by either vision-language or purely visual pathology foundation models.

\subsection{Pseudo-label quality evaluation}
\label{sec:label_quality}

\begin{table}[t]
\centering
\caption{Instance-level evaluation for assessing pseudo-label quality (CONCH).}
\label{tab:plquality}
\begingroup
\tiny
\renewcommand{\arraystretch}{0.78}
\setlength{\tabcolsep}{2.6pt}
\resizebox{\columnwidth}{!}{%
\begin{tabular}{@{}ccccc@{}}
\toprule
Dataset
& Method
& $\mathrm{AUC}_{\mathcal{Y}}$
& $\mathrm{AUC}_{\mathcal{C}}$
& $\mathrm{AUC}_{\mathcal{C}\mid \mathrm{FG}}$ \\
\midrule

\multirow{8}{*}{\shortstack[l]{TCGA-\\PanGlioma}}
& Text-Similarity
& 71.62\smallpm0.52
& 82.32\smallpm0.38
& 94.12\smallpm0.33 \\

& TextSim-Coord
& 72.93\smallpm2.35
& 80.55\smallpm2.27
& 90.35\smallpm2.33 \\

& Cover-Marginal
& 65.54\smallpm4.88
& 71.73\smallpm6.94
& 79.96\smallpm9.59 \\

& Cover-Total
& 71.43\smallpm0.51
& 82.50\smallpm0.97
& 93.96\smallpm1.71 \\

& Attention-Score
& 71.49\smallpm3.76
& 80.10\smallpm3.75
& 93.27\smallpm2.30 \\

& Attention-TopK
& 62.93\smallpm4.52
& 66.59\smallpm4.02
& 88.18\smallpm4.65 \\

& SeLa-MIL~\cite{OT}
& 71.17\smallpm3.42
& 71.53\smallpm3.72
& 59.26\smallpm9.45 \\

& INS~\cite{INS}
& 74.94\smallpm2.60
& 77.87\smallpm2.46
& 75.20\smallpm8.33 \\

& \ourscell{ProtoPur}
& \ourscell{\best{80.39\smallpm1.05}}
& \ourscell{\best{88.09\smallpm1.03}}
& \ourscell{\best{97.32\smallpm1.01}} \\

\midrule

\multirow{8}{*}{BRACS}
& Text-Similarity
& 75.06\smallpm0.39
& 74.83\smallpm0.31
& 76.98\smallpm1.11 \\

& TextSim-Coord
& 73.05\smallpm1.79
& 75.69\smallpm1.40
& 76.44\smallpm1.52 \\

& Cover-Marginal
& 59.43\smallpm6.38
& 58.17\smallpm6.44
& 58.20\smallpm7.01 \\

& Cover-Total
& 75.23\smallpm2.26
& 75.43\smallpm2.71
& 77.85\smallpm2.72 \\

& Attention-Score
& 52.77\smallpm1.20
& 48.54\smallpm1.39
& 42.70\smallpm5.46 \\

& Attention-TopK
& 67.83\smallpm1.40
& 65.80\smallpm1.47
& 70.35\smallpm1.58 \\

& SeLa-MIL~\cite{OT}
& 75.82\smallpm2.00
& 77.09\smallpm2.03
& 79.11\smallpm2.05 \\

& INS~\cite{INS}
& 73.58\smallpm2.20
& 72.84\smallpm2.64
& 75.72\smallpm3.27 \\

& \ourscell{ProtoPur}
& \ourscell{\best{77.80\smallpm1.85}}
& \ourscell{\best{77.26\smallpm2.07}}
& \ourscell{\best{79.18\smallpm2.08}} \\

\bottomrule
\end{tabular}%
}
\endgroup
\end{table}

Instead of evaluating pseudo-labels only through the final MIL performance, we used the pseudo-labels generated by each method to train the same instance classifier and then evaluated the trained classifier on the instance-level annotations. This setting isolates the effect of pseudo-label quality from differences in downstream MIL aggregation. 
The bag-consistency mask $\mathcal A_{Y_i}$ is applied to all methods, and pseudo-labels outside $\mathcal A_{Y_i}$ are excluded from instance-level training.
We evaluate TCGA-PanGlioma and BRACS using the CONCH encoder.
\paragraph{Coverage-first baselines}
Cover-Marginal and Cover-Total are two coverage-first pseudo-labeling baselines inspired by coverage maximization in submodular subset selection and active learning~\cite{submodular,Submodularity}.
Unlike ProtoPur, they select clusters by greedily increasing the coverage of target-class bags. Let $n_{i,l}$ denote the number of instances from bag $i$ assigned to cluster $V_l$, $s_i$ denote the current number of selected instances in bag $i$, and $r_i=\max(0,\tau-s_i)$ denote the remaining coverage need under the coverage threshold $\tau$. Cover-Marginal selects the cluster-class assignment that maximizes the marginal coverage gain:
$R_{\mathrm{marginal}}(l,c)=\sum_{i:Y_i=c}\min(r_i,n_{i,l})$.
Cover-Total first requires the candidate cluster to provide positive marginal coverage and then ranks feasible clusters by their total occurrence in class-$c$ bags:
$R_{\mathrm{total}}(l,c)=\sum_{i:Y_i=c}n_{i,l}$.
\paragraph{Other pseudo-labeling baselines}
We further compare ProtoPur with several representative pseudo-labeling strategies. Text-Similarity, motivated by~\cite{pathpt}, directly uses normalized patch-to-text similarities as patch-level pseudo-labels. It is included as a lightweight text-similarity baseline to assess whether text-side semantic scores alone can provide reliable prototype-level pseudo-labels. TextSim-Coord follows the same prototype selection procedure as ProtoPur, but replaces the class-wise prototype statistics with text-similarity statistics: it scores each prototype using the similarity between its prototype centroid and class-specific text embeddings. 
For the two text-based baselines, class text embeddings were generated using the official CONCH prompt template. These baselines require a paired text encoder, making them inapplicable to UNI.
Attention-Score uses a trained AB-MIL model and obtains soft pseudo-labels by min-max normalizing the attention scores of instances within each bag, an approach inherited from~\cite{weno}. Attention-TopK also relies on a trained AB-MIL model but assigns bag labels to instances with the top-$K$ attention scores for $c\in \mathcal C$ and negative pseudo-labels to instances with the $K$ smallest attention scores in each bag. In the experiments, we set $K$ to $20\%$ of the number of instances in each bag. For SeLa-MIL and INS, we use the instance pseudo-labels produced internally by the trained models.

Table~\ref{tab:plquality} reports the instance-level classification performance of different methods. ProtoPur achieves the best results on both TCGA-PanGlioma and BRACS across all three instance-level AUC metrics. On TCGA-PanGlioma, ProtoPur outperforms all baselines in $\mathrm{AUC}_{\mathcal Y}$, $\mathrm{AUC}_{\mathcal C}$, and $\mathrm{AUC}_{\mathcal C\mid FG}$, showing that the selected prototypes provide useful supervision for both overall instance classification and subtype-level discrimination. On BRACS, several attention-based and model-derived pseudo-labeling baselines degrade substantially, whereas ProtoPur remains the best, although its margin over SeLa-MIL in $\mathrm{AUC}_{\mathcal C}$ and $\mathrm{AUC}_{\mathcal C\mid FG}$ is small. This suggests that relying only on attention scores, text similarity, or model-internal pseudo-labels can be sensitive to negative instances and heterogeneous lesion patterns. In contrast, ProtoPur explicitly selects prototypes with high one-vs-rest class purity before applying the coverage constraint, which helps avoid frequent but non-specific clusters and leads to more precise subtype-aware instance supervision.

\subsection{Ablation study and hyperparameter sensitivity}
\label{sec:ablation}

\begin{table}[t]
\centering
\caption{Ablation study on TCGA-BRCA-Invasive using CLAM baseline (CONCH).}
\label{tab:ablation}
\begingroup
\scriptsize
\renewcommand{\arraystretch}{0.96}
\setlength{\tabcolsep}{3.2pt}
\resizebox{\columnwidth}{!}{%
\begin{tabular}{@{}ccccccc@{}}
\toprule
\multirow{2}{*}{Baseline}
& \multirow{2}{*}{ProtoPur}
& \multirow{2}{*}{Class-Balance}
& \multicolumn{1}{c}{Bag}
& \multicolumn{3}{c}{Instance} \\
\cmidrule(lr){4-4}\cmidrule(lr){5-7}
& & & AUC
& $\mathrm{AUC}_{\mathcal{Y}}$
& $\mathrm{AUC}_{\mathcal{C}}$
& $\mathrm{AUC}_{\mathcal{C}\mid \mathrm{FG}}$ \\
\midrule
$\checkmark$
& -- & --
& 87.23 & 83.51 & 81.15 & 77.16 \\
$\checkmark$
& $\checkmark$ & --
& 88.36 & 84.24 & 82.62 & 79.36 \\
\ourscell{$\checkmark$}
& \ourscell{$\checkmark$}
& \ourscell{$\checkmark$}
& \ourscell{\best{88.39}}
& \ourscell{\best{84.76}}
& \ourscell{\best{83.43}}
& \ourscell{\best{79.76}} \\
\bottomrule
\end{tabular}%
}
\endgroup
\end{table}

\begin{figure}[t]
\centering
\includegraphics[width=1.0\columnwidth,keepaspectratio]{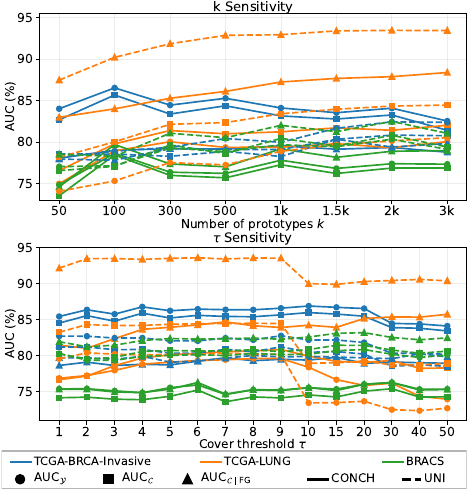}
\caption{Sensitivity of ProtoPur to the number of prototypes $k$ and the coverage threshold across three cohorts and two encoders.}
\label{fig:sensitivity}
\end{figure}

Table~\ref{tab:ablation} isolates the contribution of ProtoPur on TCGA-BRCA-Invasive using CLAM with CONCH features as the baseline. Adding ProtoPur improves all instance-level metrics, increasing $\mathrm{AUC}_{\mathcal Y}$ from 83.51 to 84.24, $\mathrm{AUC}_{\mathcal C}$ from 81.15 to 82.62, and $\mathrm{AUC}_{\mathcal C\mid FG}$ from 77.16 to 79.36, and also increases the bag-level AUC from 87.23 to 88.36. Introducing class-balanced prototype selection further improves both bag-level and instance-level performance, achieving the best results across all metrics, with $\mathrm{AUC}_{\mathcal Y}$, $\mathrm{AUC}_{\mathcal C}$, and $\mathrm{AUC}_{\mathcal C\mid FG}$ reaching 84.76, 83.43, and 79.76, respectively. These results show that purity-based pseudo-labels provide more reliable instance-level supervision, and that class-balanced selection helps reduce bias toward dominant or frequent prototype regions.

We further examine the sensitivity of ProtoPur to the prototype number $k$ and the coverage threshold $\tau$ by evaluating the pseudo-label quality as in Section~\ref{sec:label_quality}. As shown in Fig.~\ref{fig:sensitivity}, an overly small $k$ yields coarse prototypes that mix subtype-specific and shared morphology, whereas an overly large $k$ fragments the feature space into small prototypes whose purity estimates become noisy. For $\tau$, smaller values generally perform better, since a large $\tau$ forces ProtoPur to admit lower-purity prototypes to satisfy coverage. Nevertheless, the instance-level AUCs remain stable over a broad range of both $k$ and $\tau$, indicating that ProtoPur is not highly sensitive to their specific choices and supporting the default settings used in the main experiments.

\section{Discussion}
\paragraph{Identifiability and coverage} ProtoPur is designed for selective identification rather than exhaustive semantic decomposition. It controls the impurity of selected pseudo-labels under exposure assumptions, while the coverage residual captures singleton evidence not selected by a purity-first rule. If every important morphological core is required to satisfy coverage for some training bags, then ProtoPur will be forced to select those cores; otherwise, an easy core may cover all bags and cause a larger coverage residual.

\paragraph{Encoder dependency} ProtoPur is encoder-agnostic but representation-dependent. The encoder determines whether target-specific evidence forms compact and class-pure regions in the feature space, while ProtoPur uses weak bag labels to identify such regions. The main comparison shows that ProtoPur improves instance-level metrics under both CONCH and UNI features, suggesting that the method is applicable to both vision-language and vision-only pathology representations. The empirical results also suggest that the largest gains occur when the baseline already provides usable bag-level discrimination but produces noisy instance-level supervision. This is consistent with the role of ProtoPur: it does not replace representation learning, but converts class-balanced prototype purity into a more reliable instance-level training signal.

\paragraph{Decomposing the complement class}
The complement $\botcls$ groups heterogeneous instances, including shared or non-specific morphology, artifacts, and true negatives into a single class. A natural concern is that forcing such distinct patterns into one class may harm the instance classifier. From the perspective of the singleton-support formulation, $\botcls$ is an operational abstention class rather than a biological category: it only records the absence of subtype-specific evidence. To test whether a finer structure of $\botcls$ is beneficial, we split $\botcls$ into several sub-classes and retrain the instance classifier on TCGA-BRCA-Invasive with the CONCH encoder.

The subtype pseudo-labels $I_c(\tau)$, the unassigned instances, the bag-consistency mask, and all hyperparameters are identical to the default setting; only the labels of instances assigned to $\botcls$ are changed. The prototypes that are not selected by any target are grouped into $K$ sub-classes $\botcls_1,\ldots,\botcls_{k'}$ by $k$-means on their centroids, and every instance inherits the sub-class of its prototype. The instance classifier is trained with $|\mathcal C|+k'$ outputs, and $k'=1$ recovers the default single-$\botcls$ formulation. At test time, the probabilities of all $\botcls$ sub-classes are summed into a single $\botcls$ probability, so that all variants are evaluated in the same label space with the instance-level metrics of Section~\ref{sec:exp_setting}. We evaluate $k'\in\{1,2,4,8,16\}$.

\begin{table}[t]
\centering
\caption{Instance-level results with a single complement class versus decomposed complement classes on TCGA-BRCA-Invasive with the CONCH encoder.}
\label{tab:bot}
\begingroup
\scriptsize
\renewcommand{\arraystretch}{0.96}
\setlength{\tabcolsep}{3.2pt}
\resizebox{\columnwidth}{!}{%
\begin{tabular}{@{}cccc@{}}
\toprule
Complement formulation
& $\mathrm{AUC}_{\mathcal{Y}}$
& $\mathrm{AUC}_{\mathcal{C}}$
& $\mathrm{AUC}_{\mathcal{C}\mid \mathrm{FG}}$ \\
\midrule
\ourscell{Single $\botcls$ ($k'=1$, default)}
& \ourscell{84.46\smallpm1.44}
& \ourscell{83.38\smallpm1.64}
& \ourscell{79.20\smallpm3.45} \\
Decomposed $\botcls$ ($k'=2$)
& 84.14\smallpm1.53 & 83.56\smallpm2.00 & 79.47\smallpm3.53 \\
Decomposed $\botcls$ ($k'=4$)
& 84.36\smallpm1.56 & 83.61\smallpm2.20 & 79.36\smallpm3.67 \\
Decomposed $\botcls$ ($k'=8$)
& 83.49\smallpm1.87 & 82.92\smallpm2.55 & 79.05\smallpm3.36 \\
Decomposed $\botcls$ ($k'=16$)
& 84.12\smallpm1.46 & 83.62\smallpm1.95 & 79.71\smallpm2.62 \\
\bottomrule
\end{tabular}%
}
\endgroup
\end{table}

As shown in Table~\ref{tab:bot}, decomposing $\botcls$ has little effect on instance-level performance. Across $k'\in\{2,4,8,16\}$, $\mathrm{AUC}_{\mathcal Y}$ varies from 83.49 to 84.36 compared with 84.46 for the single-$\botcls$ formulation, $\mathrm{AUC}_{\mathcal C}$ from 82.92 to 83.62 compared with 83.38, and $\mathrm{AUC}_{\mathcal C\mid FG}$ from 79.05 to 79.71 compared with 79.20. All differences are at most about one point and well within the fold-to-fold standard deviations, and no value of $k'$ is consistently better across the three metrics. Note that $\mathrm{AUC}_{\mathcal C\mid FG}$ is computed on foreground instances only and therefore isolates subtype-to-subtype discrimination, whereas $\mathrm{AUC}_{\mathcal Y}$ and $\mathrm{AUC}_{\mathcal C}$ also depend on how well subtype scores are separated from non-subtype instances. These results support treating $\botcls$ as an abstention class: the classifier only needs to separate subtype-specific evidence from everything else, and the internal heterogeneity of $\botcls$ does not impair subtype discrimination. We therefore keep the simpler single-$\botcls$ formulation by default; finer decompositions remain compatible with ProtoPur, since only the labels within $\botcls$ change while the purity-first selection is unaffected.

\section{Conclusion}
We proposed ProtoPur, a prototype-purity pseudo-labeling framework for weakly supervised instance-level learning in pathology WSIs. ProtoPur estimates class-balanced local prototype purity from bag-level labels and selects high-purity prototype regions as pseudo-labeled evidence.  
We proved the singleton-support instance learnability of ProtoPur. 
Experiments demonstrated that ProtoPur consistently improves instance-level performance while maintaining strong bag-level performance. Pseudo-label quality evaluation confirmed that prototype-purity selection provides stronger instance supervision. We further discussed the identifiability of ProtoPur, its dependence on the encoder, and the effect of decomposing the complement class. These results show the advantages of ProtoPur for instance-level weakly supervised learning in pathology.

\section*{Acknowledgments}
This work was supported by the National Natural Science Foundation of China (No. 62671409) and the Sichuan Provincial Natural Science Foundation (No. 2026NSFSC1491).

\bibliographystyle{IEEEtran}
\bibliography{reference}

\end{document}